\documentclass[11pt]{article}
\usepackage{acl2012}
\usepackage{times}
\usepackage{latexsym}
\usepackage{amsmath}
\usepackage{multirow}
\usepackage{url}
\usepackage{graphicx}
\usepackage{enumitem}

\usepackage{pdfsync}
\usepackage{graphicx}
\usepackage{booktabs}
\usepackage{amssymb}
\usepackage{multicol}
\usepackage{watermark}
\usepackage{bm}

\newcommand{\toggle}[1]{}
\renewcommand{\toggle}[1]{#1} 

\newif\ifcomment\commenttrue
\ifcomment
\newcommand{\macomment}[1]{\marginpar{\begin{center}\textcolor{orange}{#1}\end{center}}}

\else
\newcommand{\macomment}[1]{} 
\fi

\def\textit#1{{\it #1}}
\def\textbf#1{{\bf #1}}
\def\textsl#1{{\sl #1}}
\def\texttt#1{{\tt #1}}

\renewcommand{\Re}{\mathbb{R}}

\title{Strategies for Training Large Vocabulary Neural Language Models}

\toggle{
\author{Wenlin Chen$^{\dag}$ \\
  Washington University\\
  St Louis, MO\\
  {\tt wenlinchen@wustl.edu} \\\And
  David Grangier\\
  Facebook AI Research\\
  Menlo Park, CA \\
  {\tt grangier@fb.com} \\\And
  Michael Auli\\
  Facebook AI Research\\
  Menlo Park, CA\\
  {\tt michaelauli@fb.com} \\}
}

\date{}

\begin{document}
\maketitle

{\let\thefootnote\relax\footnotetext{$^\dag$Work done while Wenlin was an intern at Facebook.}}

\begin{abstract}
Training neural network language models over large vocabularies is still computationally very costly compared to count-based models such as Kneser-Ney.
At the same time, neural language models are gaining popularity for many applications such as speech recognition and machine translation whose success depends on scalability.
We present a systematic comparison of strategies to represent and train large vocabularies, including softmax, hierarchical softmax, target sampling, noise contrastive estimation and self normalization. We further extend self normalization to be a proper estimator of likelihood and introduce an efficient variant of softmax. We evaluate each method on three popular benchmarks, examining performance on rare words, the speed/accuracy trade-off and complementarity to Kneser-Ney.

\end{abstract}

\section{Introduction}
\label{section:intro}

Neural network language models \cite{bengio:2003:jmlr,mikolov:2010:interspeech} have gained popularity for tasks such as automatic speech recognition \cite{arisoy:2012:wfnlm} and statistical machine translation \cite{schwenk:2012:wfnlm,vaswani:2013:emnlp}. 
Furthermore, models similar in architecture to neural language models have been proposed for translation \cite{le:2012:naacl,devlin:2014:acl,bahdanau:2015:iclr}, summarization \cite{chopra:2015:emnlp} and language generation \cite{sordoni:2015:naacl}.

Language models assign a probability to a word given a context of preceding, and possibly subsequent, words. 
The model architecture determines how the context is represented and there are several choices including recurrent neural networks \cite{mikolov:2010:interspeech}, or log-bilinear models \cite{mnih+hinton:2008:nips}. 
We experiment with a simple but proven feed-forward neural network model similar to Bengio et al. \shortcite{bengio:2003:jmlr}. 
Our focus is not the model architecture or how the context can be represented but rather how to efficiently deal with large output vocabularies, a problem common to all approaches to neural language modeling and related tasks such as machine translation and language generation.

Practical training speed for these models quickly decreases as the vocabulary grows. 
This is due to three combined factors.
First, model evaluation and gradient computation become more time consuming, mainly due to the need of computing normalized probabilities over a large vocabulary. 
Second, large vocabularies require more training data in order to observe enough instances of infrequent words which increases training times.
Third, a larger training set often allows for higher capacity models which requires more training iterations.

In this paper we provide an overview of popular strategies to model large vocabularies for language modeling. This includes the classical {\it softmax} over all output classes, {\it hierarchical softmax} which introduces latent variables, or clusters, to simplify normalization, target sampling which only considers a random subset of classes for normalization, {\it noise contrastive estimation} which discriminates between genuine data points and samples from a noise distribution, and {\it infrequent normalization}, also referred as self-normalization, which computes the partition function at an infrequent rate. We also extend self-normalization to be a proper estimator of likelihood. Furthermore, we introduce {\it differentiated softmax}, a novel variation of softmax which assigns more capacity to frequent words and which we show to be faster and more accurate than softmax (\textsection\ref{section:methods}).

Our comparison assumes a reasonable budget of one week for training models. 
We evaluate on three well known benchmarks differing in the amount of training data and vocabulary size, that is Penn Treebank, Gigaword and the recently introduced Billion Word benchmark (\textsection\ref{section:exp_setup}).

Our results show that conclusions drawn from small datasets do not always generalize to larger settings. 
For instance, hierarchical softmax is less accurate than softmax on the small vocabulary Penn Treebank task but performs best on the very large vocabulary Billion Word benchmark, because hierarchical softmax is the fastest method for training and can perform more training updates in the same period of time. Furthermore, our results with differentiated softmax  demonstrate that assigning capacity where it has the most impact allows to train better models in our time budget (\textsection\ref{section:results}).

Unlike traditional count-based models, our neural models benefit less from more training data because the computational complexity of training is much higher, exceeding our time budget in some cases.
Finally, our analysis shows clearly that Kenser-Ney count-based language models are very competitive on rare words, contrary to the common belief that neural models are better on infrequent words (\textsection\ref{section:analysis}).
\section{Modeling Large Vocabularies}
\label{section:methods}

We first introduce our basic language model architecture with a classical softmax and then describe various other methods including a novel variation of softmax.

\subsection{Softmax Neural Language Model}
\label{section:softmax}

Our feed-forward neural network implements an n-gram language model, i.e., it is a parametric function estimating the probability of the next word $w^t$ given $n-1$ previous context words, $w^{t-1},\ldots, w^{t-n+1}$. 
Formally, we take as input a sequence of discrete indexes representing the $n-1$ previous words and output a vocabulary-sized vector of probability estimates, i.e.,
$$
f : \{1,\ldots, V\}^{n-1} \to [0,1]^V,
$$
where $V$ is the vocabulary size. 
This function results from the composition of simple differentiable functions or {\it layers}. 

Specifically, $f$ composes an input mapping from discrete word indexes to continuous vectors, a succession of linear operations followed by hyperbolic tangent non-linearities, plus one final linear operation, followed by a softmax normalization. 

The input layer maps each context word index to a continuous $d_0$-dimensional vector. It relies on a parameter matrix $W^0 \in \Re^{V \times d_0}$ to convert the input 
$$
x = [w^{t-1},\ldots, w^{t-n+1}] \in \{1,\ldots, V\}^{n-1}
$$
to $n-1$ vectors of dimension $d_0$. These vectors are concatenated into a single $(n-1) \times d_0$ matrix,
$$
h^0 = [W^0_{w^{t-1}}; \ldots; W^0_{w^{t-n+1}} ] \in \Re^{n-1 \times d_0}.
$$
This state $h^0$ is considered as a $(n-1) \times d_0$ vector by
the next layer. The subsequent states are computed through $k$ layers of linear mappings followed by hyperbolic tangents, i.e.
$$
\forall i = 1, \ldots, k, \quad h^{i} = {\rm tanh}(W^i h^{i-1} + b^i) \in \Re^{d_i}
$$
where $W^i \in \Re^{d_i \times d_{i-1}}, b \in \Re^{d_i}$ are learnable weights and biases and ${\rm tanh}$ denotes the component-wise hyperbolic tangent.

Finally, the last layer performs a linear operation followed by a softmax normalization, i.e.,
\begin{equation}\label{equation:softmax}
h^{k+1} = W^{k+1} h^{k} + b^{k+1} \in \Re^{V}
\end{equation}
\begin{align}
\label{equation:softmax_z}
\textrm{and}\quad
&y = \frac{1}{Z} ~ \exp(h^{k+1}) \in [0,1]^V \\\textrm{where}\quad\nonumber 
&Z = \sum_{j=1}^V \exp(h^{k+1}_j).
\end{align}
and $\exp$ denotes the component-wise exponential. 
The network output $y$ is therefore a vocabulary-sized vector of probability estimates.
We use the standard cross-entropy loss with respect to the computed log probabilities
$$
\frac{\partial \log y_i}{\partial h^{k+1}_j} = \delta_{ij} - y_j
$$
where $\delta_{ij} = 1$ if $i = j$ and $0$ otherwise
The gradient update therefore increases the score of the correct output $h^{k+1}_i$ and decreases the score of all other outputs $h^{k+1}_j$ for $j \ne i$.

A downside of the classical softmax formulation is that it requires computation of the activations for {\it all output words} (see Equation~\ref{equation:softmax_z}). 
When grouping multiple input examples into a batch, Equation~\ref{equation:softmax} amounts to a large matrix-matrix product of the form $W^{k+1}H^{k}$ where $W^{k+1} \in \Re^{V \times d_k}$, $H^k = [h^k_1; \ldots; h^k_l] \in \Re^{d_k \times l}$, where $l$ is the number of input examples in a batch.
For example, typical settings for the gigaword corpus (\textsection\ref{section:exp_setup}) are a vocabulary of size $V = 100,000$, with output word embedding size $d_k = 1024$ and batch size of $l=500$ examples.
This gives a {\it very large} matrix-matrix product of $100,000 \times 1024$ by $1024 \times 500$. The rest of the network involves matrix-matrix operations whose size is determined by the batch size and the layer dimensions, both are typically much smaller than the vocabulary size, ranging for hundreds to a couple of thousands.
Therefore, the output layer dominates the complexity of the entire network.

This computational burden is high even for Graphics Processing Units (GPUs). 
GPUs are well suited for matrix-matrix operation when matrix dimensions are 
in the thousands, but become less efficient with dimensions over $10,000$. 
The size of the output matrix is therefore a bottleneck during training.
Previous work suggested tackling these products by sharding them across multiple GPUs \cite{sutskever:2014:nips}, which introduces additional engineering challenges around inter-GPU communication. This paper focuses on orthogonal algorithmic solutions which are also relevant to parallel training.

\subsection{Hierarchical Softmax}
\label{section:hsm}

Hierarchical Softmax (HSM) organizes the output vocabulary into a tree where the leaves are the words and the intermediate nodes are latent variables, or {\it classes} \cite{morin:2005:aistats}.
The tree has potentially many levels and there is a unique path from the root to each word.
The probability of a word is the product of the probabilities of the latent variables along the path from the root to the leaf, including the probability of the leaf.
If the tree is perfectly balanced, this can reduce the complexity from $\mathcal{O}(V)$ to $\mathcal{O}(\log V)$.

We experiment with a version that follows Goodman \shortcite{goodman:2001:icassp} and which has been used in Mikolov et al. \shortcite{mikolov:2011:icassp}.
Goodman proposed a two-level tree which first predicts the {\it class} of the next word $c^t$ and then the actual word $w^t$ given context $x$
\begin{equation}
p(w^t|x) = p(c^t|x)~p(w^t|c^t, x)\label{eq:hsm}
\end{equation}
If the number of classes is $\mathcal{O}(\sqrt{V})$ and each class has the same number of members, then we only need to compute $\mathcal{O}(2~\sqrt{V})$ outputs. This is a good strategy in practice as it yields weight matrices for clusters and words whose largest dimension is less than $\sim 1,000$, a setting for which GPUs are fast.

A popular strategy clusters words based on {\it frequency}. It slices the list of words sorted by frequency into clusters that contain an equal share of the total unigram probability.
We pursue this strategy and compare it to random class assignment and to clustering based on word embedding features. The latter applies k-means over word embeddings obtained from Hellinger PCA over co-occurrence counts~\cite{lebret:2014:eacl}.  
Alternative word representations~\cite{brown:1992:cl,mikolov:2013:corr} are also relevant but an extensive study of word clustering techniques is beyond the scope of this work.

\subsection{Differentiated Softmax}
\label{section:differentiated_softmax}
This section introduces a novel variation of softmax that assigns variable capacity per word in the output layer. The weight matrix of the final layer $W^{k+1} \in \Re^{d_k \times V}$ stores {\it output embeddings} of size $d_k$ for the $V$ words the language model may predict: $W^{k+1}_1; \ldots; W^{k+1}_V$.
Differentiated softmax (D-Softmax) varies the dimension of the output embeddings $d_k$ across words depending on how much model capacity is deemed suitable for a given word.
In particular, it is meaningful to assign more parameters to frequent words than to rare words. By definition, frequent words occur more of ten in the training data than rare words and therefore allow to fit more parameters.

In particular, we define partitions of the output vocabulary based on word frequency and the words in each partition share the same embedding size.
For example, we may partition the frequency ordered set of output word ids, $O = \{1, \ldots, V\}$, into $A^{d_A} = \{1, \ldots, K\}$ and $B^{d_B} = \{K+1, \ldots, V\}~$ s.t. $~A \cup B = O~\wedge~A \cap B = \emptyset$, where $d_A$ and $d_B$ are different output embedding sizes and $K$ is a word id.

\begin{figure}
\centering
\includegraphics[width=0.25\textwidth]{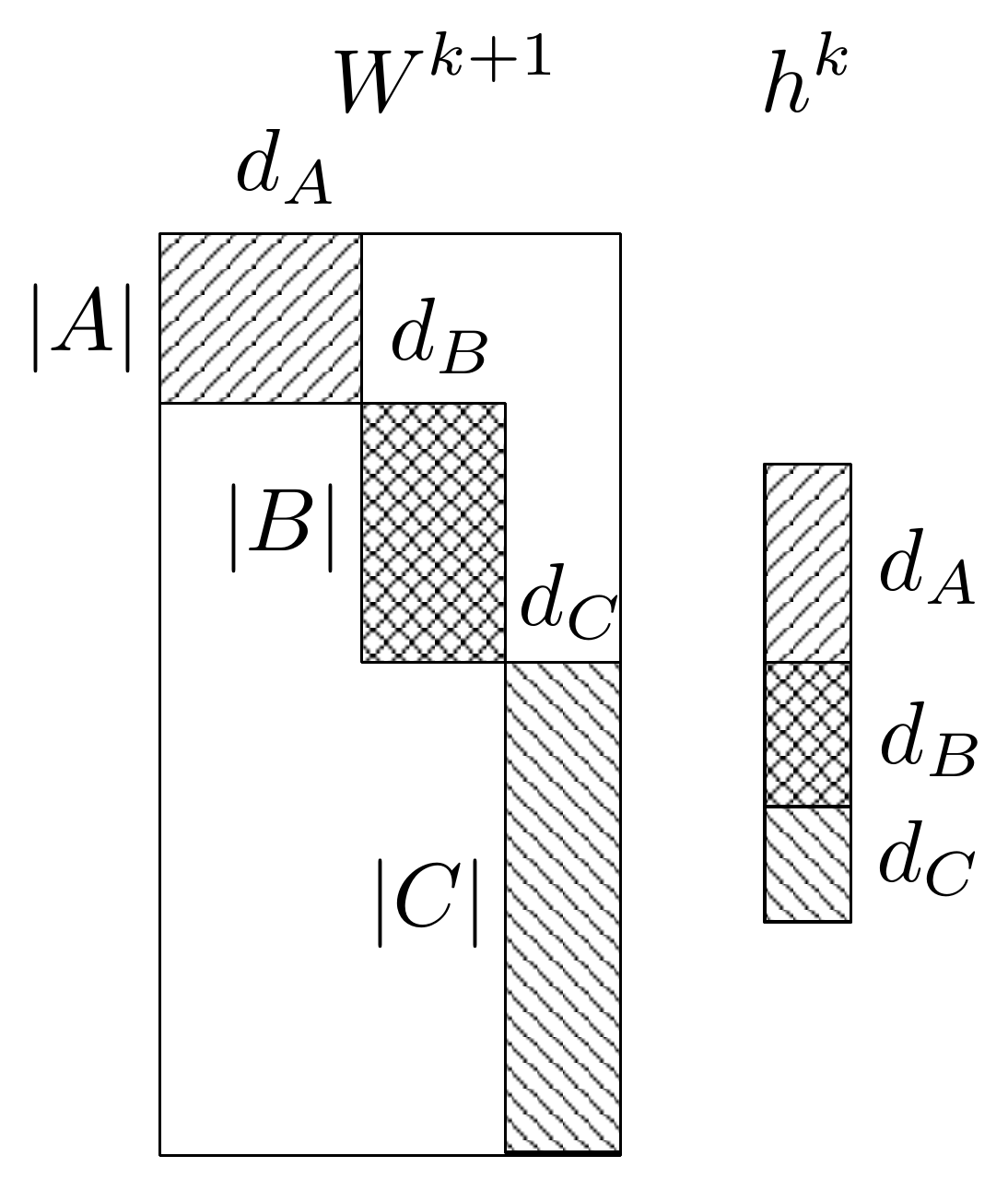}
\caption{Final weight matrix $W^{k+1}$ and hidden layer $h^k$ for differentiated softmax for partitions $A, B, C$ of the output vocabulary with embedding dimensions $d_A, d_B, d_C$; non-shaded areas are zero.}
\label{figure:dfsm_weights}
\end{figure}

Partitioning results in a sparse final weight matrix $W^{k+1}$ which arranges the embeddings of the output words in blocks, each one corresponding to a separate partition (Figure~\ref{figure:dfsm_weights}).
The size of the final hidden layer $h^k$ is the sum of the embedding sizes of the partitions.  The final hidden layer is effectively a concatenation of separate features for each partition which are used to compute the dot product with the corresponding embedding type in $W^{k+1}$.
In practice, we compute separate matrix-vector products, or in batched form, matrix-matrix products, for each partition in $W^{k+1}$ and $h^k$.

Overall, differentiated softmax can lead to large speed-ups as well as accuracy gains since we can greatly reduce the complexity of computing the output layer.
Most significantly, this strategy speeds up {\it both} training and inference. 
This is in contrast to hierarchical softmax which is fast during training but requires even more effort than softmax for computing the most likely next word.

\subsection{Target Sampling}
\label{section:target_sampling}

Sampling-based methods approximate the softmax normalization (Equation~\ref{equation:softmax_z}) by selecting a number of {\it impostors} instead of using all outputs. This can significantly speed-up each training iteration, depending on the size of the impostor set.

We follow Jean et al. \shortcite{jean:2014:arxiv} who choose as impostors all positive examples in a mini-batch as well as a subset of the remaining words. This subset is sampled uniformly and its size is chosen by cross-validation. 
A downside of sampling is that the (downsampled) final weight matrix $W^{k+1}$ (Equation~\ref{equation:softmax}) keeps changing between mini-batches. 
This is computationally costly and the success of sampling hinges on being to estimate a good model while keeping the number of samples small.

\subsection{Noise Contrastive Estimation}
\label{section:nce}

Noise contrastive estimation (NCE) is another sampling-based technique~\cite{gutmann:2010:aistats,mnih:2012:icml}. Contrary to target sampling, it does not maximize the training data likelihood directly. 
Instead, it solves a two-class problem of distinguishing genuine data from noise samples. 
The training algorithm samples a word $w$ given the preceding context $x$ from a mixture
$$
P(w|x) = \frac{1}{k+1} P_{\rm train}(w|x) + \frac{k}{k+1} P_{\rm noise}(w|x)
$$
where $ P_{\rm train}$ is the empirical distribution of the training set and $P_{\rm noise}$ is a known noise distribution which is typically a context-independent unigram distribution fitted on the training set. 
The training algorithm fits the model $\hat{P}(w|x)$ to recover whether a mixture sample came from the data or the noise distribution, this amounts to minimizing the binary cross-entropy
$$
-y ~\log \hat{P}(y = 1| w, x) - (1-y) ~\log \hat{P}(y = 0 | w, x)
$$
where $y$ is a binary variable indicating whether the current sample originates
from the data ($y=1$) or the noise ($y=0$) and 
$
\hat{P}(y = 1| w, x) = \frac{\hat{P}(w|x)}{\hat{P}(w|x)+k P_{\rm noise}(w|x)},
$
$
\hat{P}(y = 0| w, x) = 1 - \hat{P}(y = 1| w, x)
$
are the model estimates of the corresponding posteriors.

This formulation still involves a softmax over the vocabulary to compute $\hat{P}(w|x)$. 
However, Mnih and Teh~\shortcite{mnih:2012:icml} suggest to forego the normalization step and simply consider replacing $\hat{P}(w|x)$ with unnormalized exponentiated scores which makes the complexity of training independent of the vocabulary size. 
At test time, the softmax normalization is reintroduced to obtain a proper distribution.

\subsection{Infrequent Normalization}
\label{section:weaknorm}

Andreas and Klein~\shortcite{andreas:2015:naacl} also propose to relax score normalization. 
Their strategy (here referred to as WeaknormSQ) associates unnormalized likelihood maximization with a penalty term that favors normalized predictions. This yields 
the following loss over the training set $T$
$$
L^{(2)}_{\alpha} = 
-\sum_{(w,x) \in T} s(w|x) 
+ \alpha \sum_{(w,x) \in T} (\log Z(x))^2$$
where $s(w|x)$ refers to the unnormalized score of word $w$ given context $x$ and $Z(x) = \sum_w \exp( s(w|x) )$ refers to the partition function for context $x$. 
For efficient training, the second term can be down-sampled
$$
L^{(2)}_{\alpha,\gamma} = 
-\sum_{\substack{(w,x)\\~~~\in {\rm train}}} s(w|x) 
+ \frac{\alpha}{\gamma} \sum_{\substack{(w,x)\\~~~\in {\rm train_\gamma}}} (\log Z(x))^2
$$
where $T_\gamma$ is the training set sampled at rate $\gamma$.
A small rate implies computing the partition function only for a small fraction of the training data.

This work extends this strategy to the case where the log partition term is not squared (Weaknorm), i.e.,
$$
L^{(1)}_{\alpha,\gamma} = 
-\sum_{\substack{(w,x)\\~~~\in {\rm train}}} s(w|x) 
+ \frac{\alpha}{\gamma} \sum_{\substack{(w,x)\\~~~\in {\rm train_\gamma}}} \log Z(x)$$
For $\alpha = 1$, this loss is an unbiased  estimator of the negative 
log-likelihood of the training data
$
L^{(2)}_1 = -\sum_{(w,x) \in {\rm train}} s(w|x) - \log Z(x)
$.

\subsection{Other Methods}
\label{section:other_methods}

Fast locality-sensitive hashing has been used to approximate the dot-product between the final hidden layer activation $h^k$ and the output word embedding \cite{vijayanarasimhan:2014:arxiv}. 
However, during training, there is a high overhead for re-indexing the embeddings 
and test time speed-ups virtually vanish as the batch size increases due to the 
efficiency of matrix-matrix products.
\section{Experimental Setup}
\label{section:exp_setup}

This section describes the data used in our experiments, our evaluation methodology and our validation procedure. 

\begin{description}[leftmargin=0pt]

\item[Datasets] Our experiments are performed over three datasets of different sizes: Penn Treebank (PTB), WMT11-lm (billionW) and English Gigaword, version 5 (gigaword). 
Penn Treebank is a well-established dataset for evaluating language models~\cite{marcus:1993:cl}. 
It is the smallest dataset with a benchmark setting relying on 1 million tokens and a vocabulary size of $10,000$ \cite{mikolov:2011:interspeech}. 
The vocabulary roughly corresponds to words occurring at least twice in the training set.
The WMT11-lm corpus has been recently introduced as a larger corpus to evaluate language models and their impact on statistical machine translation~\cite{chelba:2013:techreport}. 
It contains close to a billion tokens and a vocabulary of about 800,000 words, which corresponds to words with more than $3$ occurrences in the training set.\footnote{We use the version distributed by Tony Robinson at \texttt{http://tiny.cc/1billionLM}~.}
This dataset is often referred as the {\it billion word benchmark}.
Gigaword~\cite{parker:2011:techreport} is the largest corpus we consider with 5 billion tokens of newswire data.
Even though it has been used for language modeling previously~\cite{heafield:2011:wmt}, there is no standard train/test split or vocabulary for this set. 
We split the data according to time: the training set covers the period 1994--2009 and the test data covers 2010. The vocabulary consists of the $100,000$ most frequent words, which roughly corresponds to words with more than $100$ occurrences in the training data. 
Table~\ref{table:datasets} summarizes data set statistics.

\begin{table}
\centering
\begin{tabular}{l r r r r}
 Dataset & Train & Test & Vocab & OOV \\ \hline
 PTB      &  1M         &     0.08M   &    10k     & 5.8\%    \\ 
 gigaword     & 4,631M      &   279M      &   100k  & 5.6\%\\
 billionW & 799M        &    8.1M     &     793k   & 0.3\%\\
\hline
\end{tabular}
\caption{Dataset statistics.  Number of tokens for train and test set, vocabulary size and ratio of out-of-vocabulary words in the test set.}
\label{table:datasets}
\end{table}

\item[Evaluation] 

Performance is evaluated in terms of perplexity over the test set. 
For PTB and billionW, we report perplexity results on a per sentence basis, i.e., the model does not use context words across sentence boundaries and we score the end-of-sentence marker. 
This is the standard setting for these benchmarks. 
On gigaword, we do not segment the data into sentences and the model uses contexts crossing sentence boundaries and the evaluation does not include end-of-sentence markers.

Our baseline is an interpolated Kneser-Ney (KN) language model and we use the KenLM toolkit to train 5-gram models without pruning~\cite{heafield:2011:wmt}. For our neural models, we train 11-gram language models for gigaword, billionW and a 6-gram language model for the smaller PTB.
The parameters of the models are the weights $W^i$ and the biases $b^i$ for $i = 0, \ldots, k+1$. 
These parameters are learned by maximizing the log-likelihood of the training data relying on stochastic gradient descent (SGD)~\cite{lecun:1998:tricks}.

\item[Validation]
The hyper-parameters of the model are the number of layers $k$ and the dimension of each layer $d_i, \forall i = 0, \ldots, k$.
These parameters are set by cross-validation, i.e., the parameters which maximize the likelihood over a validation set (subset of the training data excluded from sampling during SGD optimization).
We also cross-validate the number of clusters and as well as the clustering technique for hierarchical softmax, the number of frequency bands and their allocated capacity for differentiated softmax, the number of distractors for target sampling, the noise/data ratio for NCE, as well as the regularization rate and strength for infrequent normalization.
Similarly, the SGD parameters, i.e., learning rate and mini-batch size, are also set to maximize validation accuracy.

\item[Training Time] We train for 168 hours (one week) on the large datasets (billionW, gigaword) and 24 hours (one day) for Penn Treebank. We select the hyper-parameters which yield the best validation perplexity after the allocated time and report the perplexity of the resulting model on the test set. This training time is a trade-off between being able to do a comprehensive exploration of the various settings for each method and good accuracy. 
%
\end{description}

\section{Results}
\label{section:results}

\begin{figure}
\centering
\includegraphics[width=0.5\textwidth]{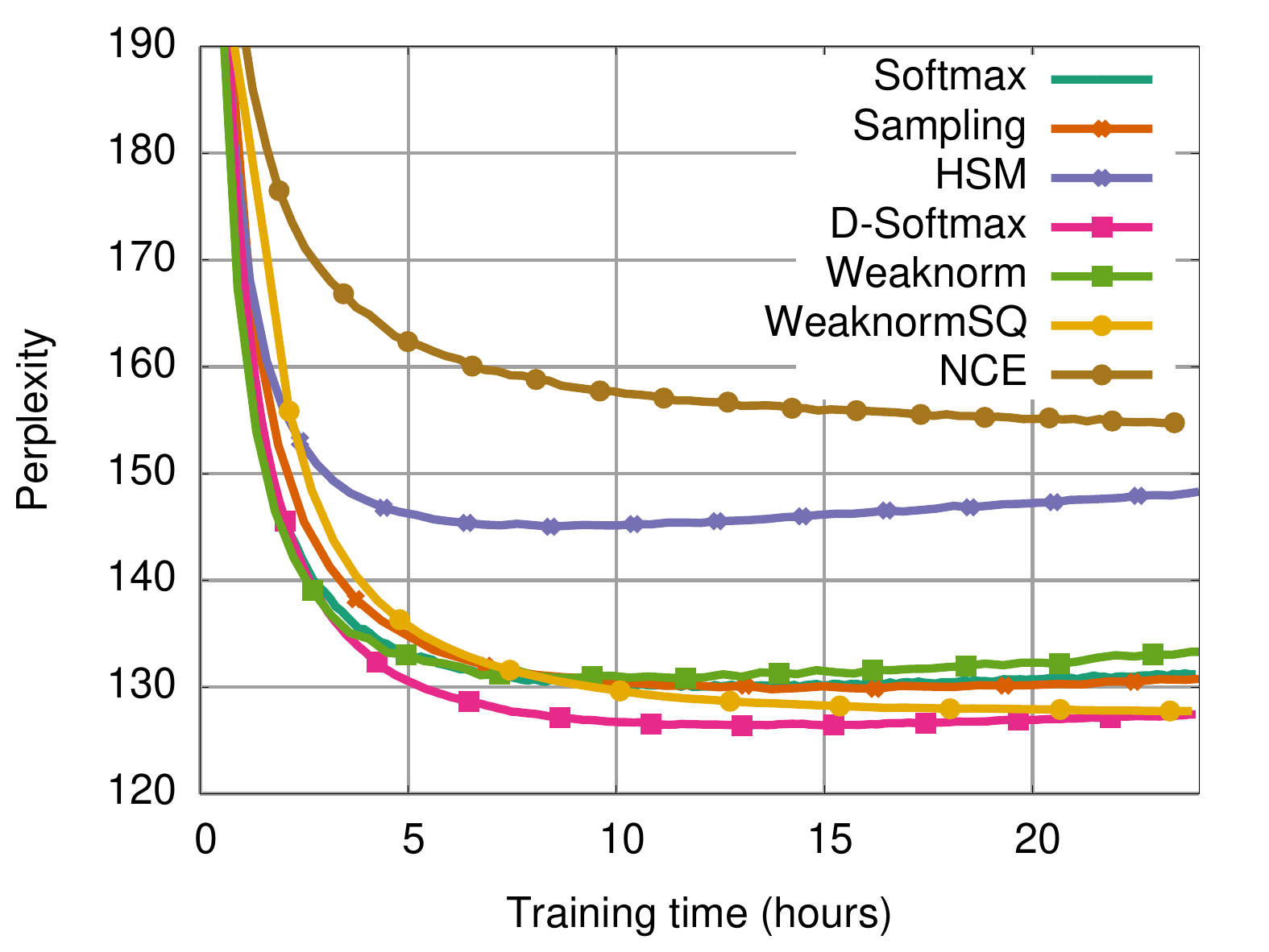}
\caption{Penn Treebank learning curve on the validation set.}
\label{figure:overview_ptb}
\end{figure}

\begin{figure}
\centering
\includegraphics[width=0.5\textwidth]{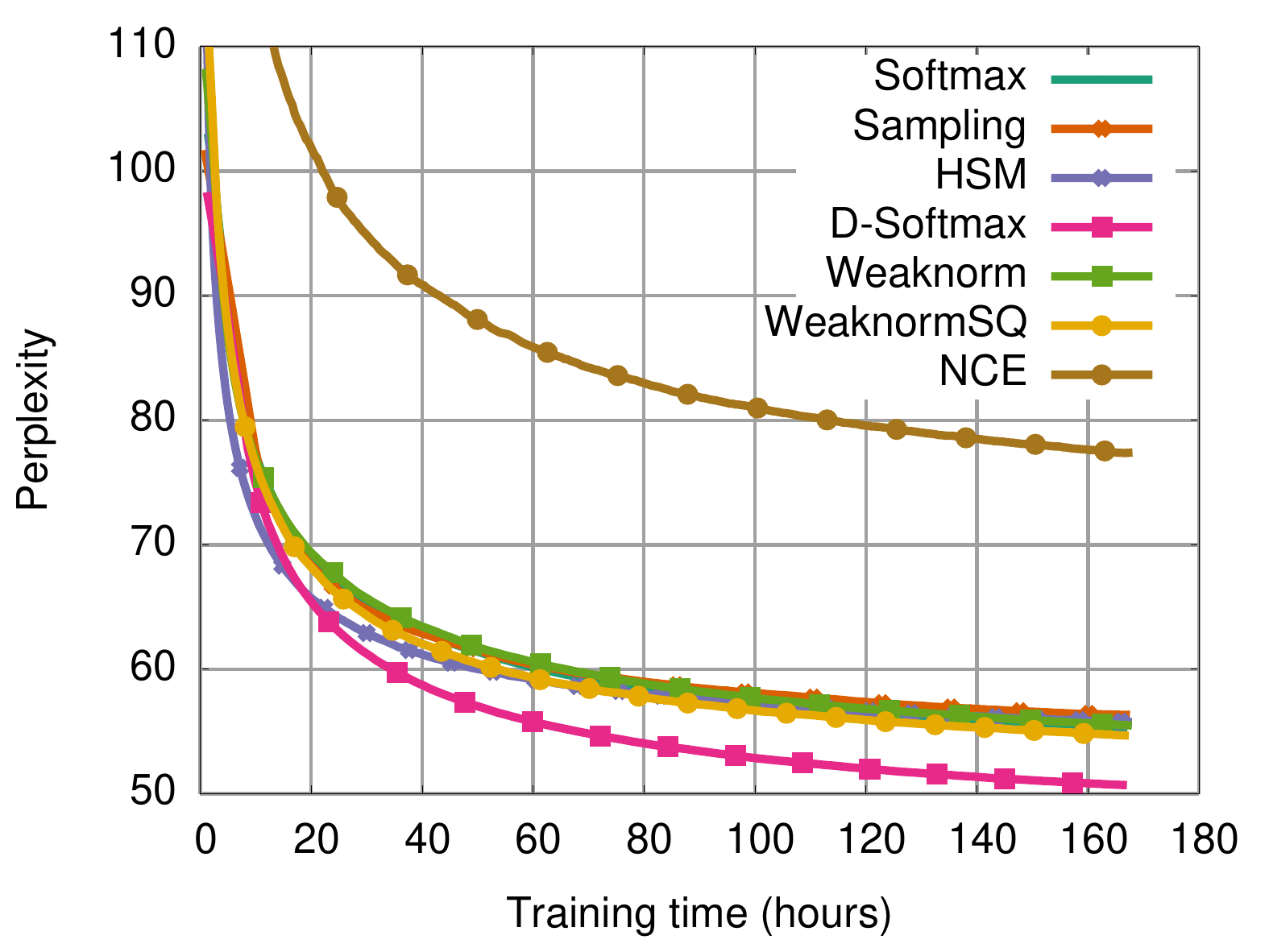}
\caption{Gigaword learning curve on the validation set.}
\label{figure:overview_giga}
\end{figure}

\begin{figure}
\centering
\includegraphics[width=0.5\textwidth]{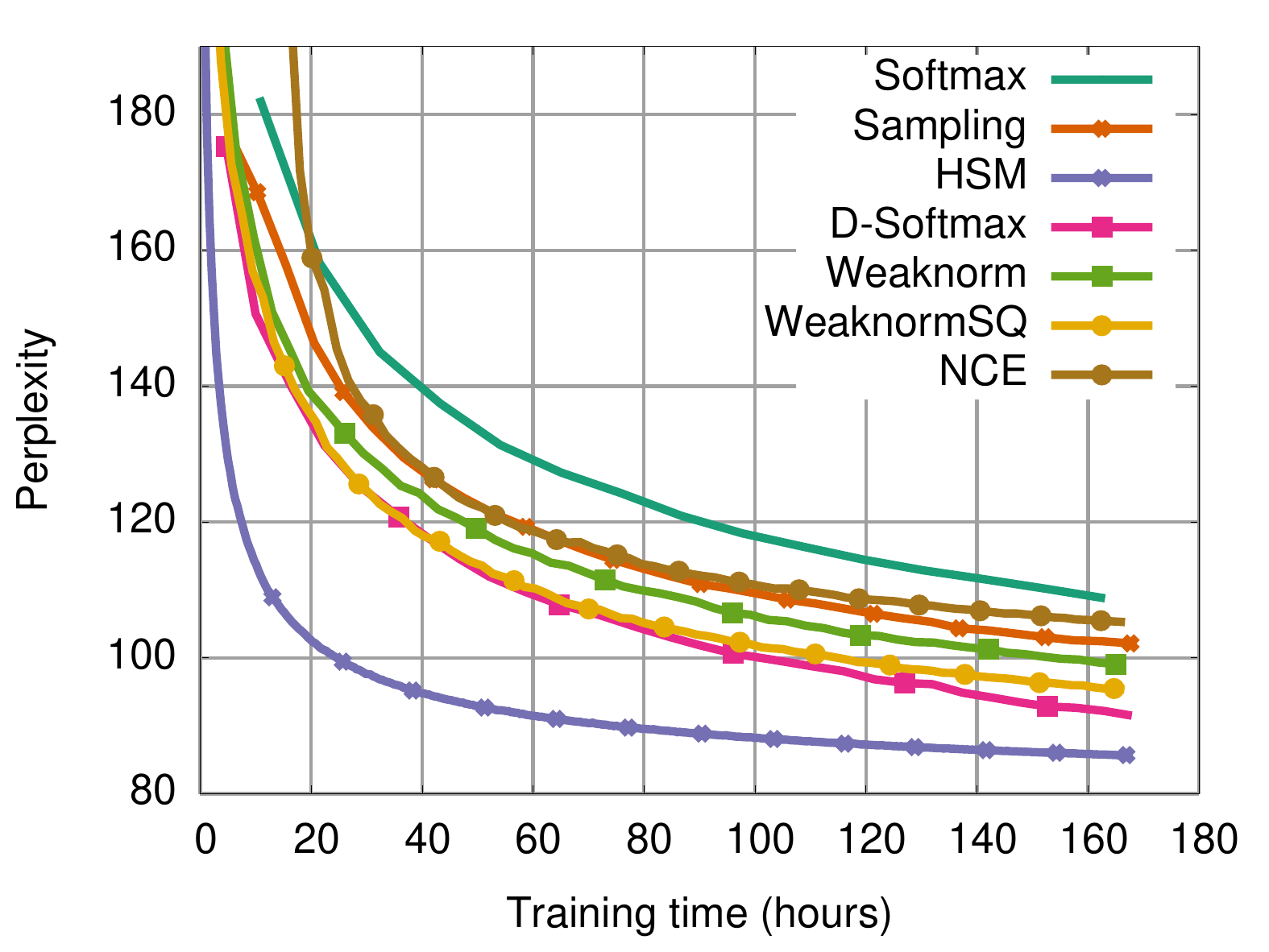}
\caption{Billion Word learning curve on the validation set.}
\label{figure:overview_billionW}
\end{figure}

Looking at test results (Table~\ref{table:overview_results}) and learning paths on the validation sets (Figures~\ref{figure:overview_ptb}, \ref{figure:overview_giga}, and \ref{figure:overview_billionW}) we can see a clear trend: the competitiveness of softmax diminishes with the vocabulary size.
Softmax does very well on the small vocabulary Penn Treebank corpus, but it does very poorly on the larger vocabulary billionW corpus.
Faster methods such as sampling, hierarchical softmax, and infrequent normalization (Weaknorm and WeaknormSQ) are much better in the large-vocabulary setting of billionW.

D-Softmax is performing very well on all data sets and shows that assigning higher capacity where it benefits most results in better models. 
Target sampling performs worse than softmax on gigaword but better on billionW.
Hierarchical softmax performs very poorly on Penn Treebank which is in stark contrast to billionW where it does very well.
Noise contrastive estimation has good accuracy on billionW, where speed is essential to achieving good accuracy.

Of all the methods, hierarchical softmax processes most training examples in a given time frame (Table~\ref{table:overview_timing}).
Our test time speed comparison assumes that we would like to find the highest scoring next word, instead rescoring an existing string.  
This scenario requires scoring all output words and D-Softmax can process nearly twice as many tokens per second than the other methods whose complexity is then similar to softmax.

\begin{table}
\centering
\begin{tabular}{l r r r}
			&	PTB	&	gigaword	&	billionW	\\\hline
KN			&	141.2	&	57.1	&	70.2	\\\hline
Softmax		&	123.8	&	56.5	&	108.3	\\
D-Softmax	&	121.1	&	52.0	&	91.2	\\
Sampling	&	124.2	&	57.6	&	101.0	\\
HSM			&	138.2	&	57.1	&	85.2	\\
NCE			&	143.1	&	78.4 	&	104.7	\\
Weaknorm	&	124.4	&	56.9	&	98.7	\\
WeaknormSQ	&	122.1	&	56.1	&	94.9	\\\hline
KN+Softmax	&	108.5	&	43.6	&	59.4	\\
KN+D-Softmax&	107.0	&	42.0	&	56.3	\\
KN+Sampling	&	109.4	&	43.8	&	58.1	\\
KN+HSM		&	115.0	&	43.9	&	55.6	\\
KN+NCE		&	114.6	&	49.0    &	58.8	\\
KN+Weaknorm	&	109.2	&	43.8	&	58.1	\\
KN+WeaknormSQ&	108.8	&	43.8	&	57.7	\\\hline

\end{tabular}
\caption{Test perplexity of individual models and interpolation with Kneser-Ney.}
\label{table:overview_results}
\end{table}

\begin{table}
\centering
\begin{tabular}{l r r}
            &        train &      test \\\hline
Softmax		&	       510 &      510	\\
D-Softmax	&	       960 & {\bf 960}	\\
Sampling	&        1,060 &      510	\\
HSM			& {\bf 12,650} &      510 \\
NCE			&	     4,520 &      510 \\
Weaknorm	&	     1,680 &      510 \\
WeaknormSQ	&	     2,870 &      510 \\\hline
\end{tabular}
\caption{Training and testing speed on billionW in tokens per second. 
Most techniques are identical to softmax at test time, HSM can be faster at test time if only few words involving few clusters are being scored.}
\label{table:overview_timing}
\end{table}

\subsection{Softmax}
\label{section:softmax_results}

Despite being our baseline, softmax ranks among the most accurate methods on PTB and it is second best on gigaword after D-Softmax (with WeaknormSQ performing similarly). For billionW, the extremely large vocabulary makes softmax training too slow to compete with faster alternatives. However, of all the methods softmax has the simplest implementation and it has no additional hyper-parameters compared to other methods.

\subsection{Target Sampling}
\label{section:target_sampling_results}

\begin{figure}
\centering
\includegraphics[width=0.5\textwidth]{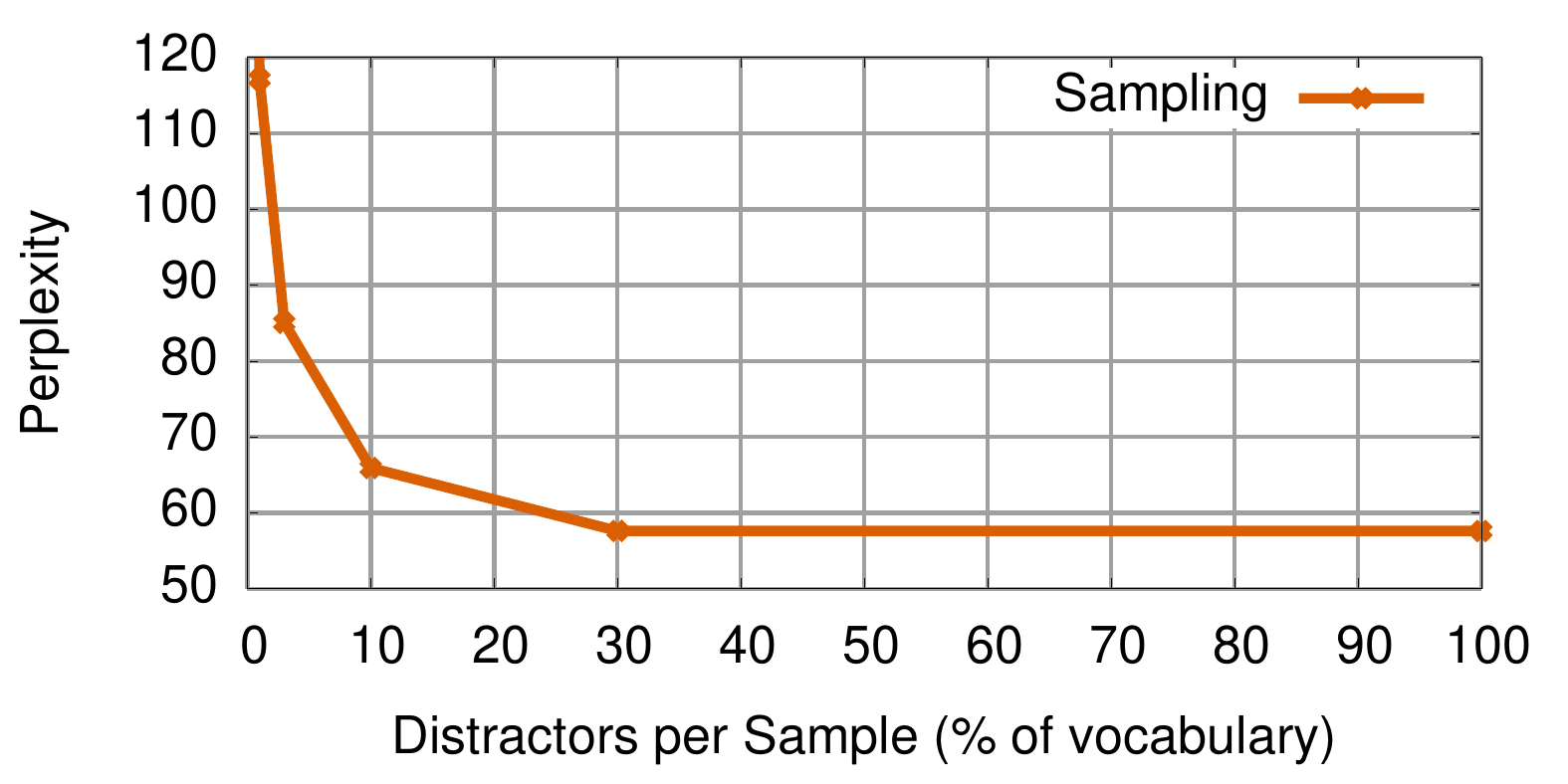}
\caption{Number of Distractors versus Perplexity for Target Sampling over Gigaword}
\label{figure:target_sampling_num_distractors}
\end{figure}

Figure~\ref{figure:target_sampling_num_distractors} shows that target sampling is most accurate when the distractor set represents a large fraction of the vocabulary, i.e. more than 30\% on gigaword (billionW best setting is even higher with 50\%). Target sampling is asymptotically faster and therefore performs more iterations than softmax in the same time. However, it makes less progress in terms of perplexity reduction per iteration compared to softmax. Overall, it is not much better than softmax. A reason might be that the sampling procedure chooses distractors independently from context, or current model performance. This does not favor sampling distractors the model incorrectly considers likely given the current context. These distractors would yield high gradient that could make the model progress faster.

\subsection{Hierarchical Softmax}
\label{section:hsm_results}

Hierarchical softmax is very efficient for large vocabularies and it is the best method on billionW. On the other hand, HSM is performing poorly on small vocabularies as seen on Penn Treebank.

We found that a good word clustering structure helps learning: when each cluster contains words occurring in similar contexts, cluster likelihoods are easier to learn; when the cluster structure is uninformative, cluster likelihoods converge to the uniform distribution. 
This adversely affects accuracy since words can never have higher probability than their clusters (cf. Equation~\ref{eq:hsm}).

Our experiments group words into a two level hierarchy and compare four clustering strategies over billionW and gigaword (\textsection\ref{section:hsm}).
Random clustering shuffles the vocabulary and splits it into equally sized partitions.
Frequency-based clustering first orders words based on the number of their occurrences and assigns words to clusters such that each cluster represents an equal share of frequency counts~\cite{mikolov:2011:icassp}. K-means runs the well-know clustering algorithm on
Hellinger PCA word embeddings. Weighted k-means is similar but weights each word by its frequency.

Random clustering performs worst (Table~\ref{table:hsm_perplexity}) followed by frequency-based clustering but k-means does best; weighted k-means performs similarly than its unweighted version.
In our initial experiments, pure k-means performed very poorly because the most significant cluster captured up to $40\%$ of the word frequencies in the data.
We resorted to explicitly capping the frequency-budget of each cluster to $\sim 10\%$ which brought k-means to the performance of weighted k-means.


\begin{table}
\centering
\begin{tabular}{l r r r}
    			 	& billionW 	& gigaword \\\hline
random     			& 98.51		& 62,27	\\
frequency-based		& 92.02		& 59.47 \\
k-means    			& 85.70		& 57.52	\\
weighted k-means	& 85.24		& 57.09	\\
\hline
\end{tabular}
\caption{Comparison of clustering techniques for hierarchical softmax.}
\label{table:hsm_perplexity}
\end{table}

\subsection{Differentiated Softmax}

D-Softmax is the best technique on gigaword, and the second best on billionW, after HSM. On PTB it ranks among the best techniques whose perplexities cannot be reliably distinguished. The variable-capacity scheme of D-Softmax can assign large embeddings to frequent words, while keeping computational complexity manageable through small embeddings for rare words. 

Unlike for hierarchical softmax, NCE or Weaknorm, the computational advantage of D-Softmax is preserved at test time (Table~\ref{table:overview_timing}). D-Softmax is the fastest technique at test time, while ranking among the most accurate methods. This speed advantage is due to the low dimensional representation of rare words which negatively affects the model accuracy on these words (Table~\ref{table:overview_entropy}).

\subsection{Noise Contrastive Estimation}
\label{section:nce_results}

For language modeling we found NCE difficult to use in practice. In order to work with large neural networks and large vocabularies, we had to dissociate the number of noise samples from the data to noise ratio in the modeled mixture. For instance, a data/noise ratio of $1/50$ gives good performance in our experiments but estimating only $50$ noise sample posteriors per data point is wasteful given the cost of network evaluation. Moreover, this setting does not allow frequent sampling of every word in a large vocabulary. Our setting considers more noise samples and up-weights the data sample. This allows to set the data/noise ratio independently from the number of noise samples.  

\begin{figure}
\centering
\includegraphics[width=0.5\textwidth]{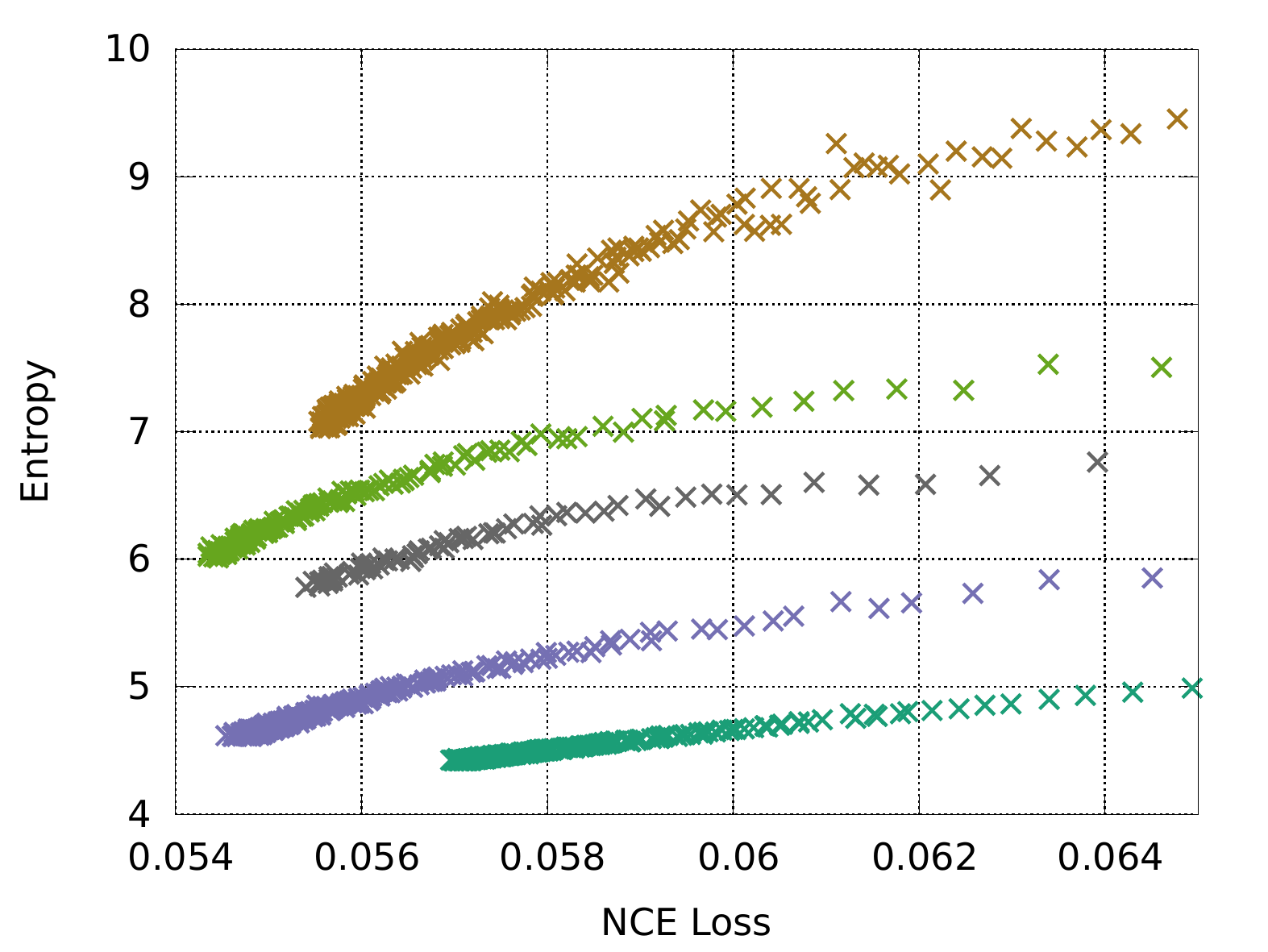}
\caption{Validation entropy versus NCE loss over gigaword for different experiments differing only in their learning rates and initial weights.}
\label{figure:scatter_nce}
\end{figure}

Overall, NCE results are better than softmax only for billionW, a setting for which softmax is very slow due to the very large vocabulary. Why does NCE perform so poorly? Figure~\ref{figure:scatter_nce} shows entropy on the validation set versus the NCE loss for several models. The results clearly show that similar NCE loss values can result in very different validation entropy. 
Although NCE might make sense for other metrics, it is not among the best techniques for minimizing perplexity.

\subsection{Infrequent Normalization}
\label{section:infrequent_normalization_results}

Infrequent normalization (Weaknorm and WeaknormSQ) performs better than softmax on billionW and comparably to softmax on Penn Treebank and gigaword (Table~\ref{table:overview_results}).
The speedup from skipping partition function computations is substantial.
For instance, WeaknormSQ on billionW evaluates the partition only on $10\%$ of the examples.
In one week, the model is evaluated and updated on 868M tokens (with 86.8M partition evaluations) compared to 156M tokens for softmax. 

Although referred to as self-normalizing in the literature~\cite{andreas:2015:naacl}, the trained models still needs to be normalized after training.
The partition cannot be considered as a constant and varies greatly between data samples. 
On billionW, the 10th to 90th percentile range was $9.4$ to $10.3$ on the natural log scale, i.e., a ratio of $2.5$ for WeaknormSQ. 

It is worth noting that the squared regularizer version of infrequent normalization (WeaknormSQ) is highly sensitive to the regularization parameter. 
We often found regularization strength to be either too low (collapse) or too high (blow-up) after a few days of training. 
We added an extra unit to our model in order to bound predictions, which yields more stable training and better generalization performance. We bounded unnormalized predictions within the range $[-10, +10]$ by using $x \to 10~\mathrm{tanh}(x/5)$). 
We also observed that for the non-squared version of the technique (Weaknorm), a regularization strength of $1$ was the best setting. With this choice, the loss is an unbiased estimator of the data likelihood.

\section{Analysis}
\label{section:analysis}

This section discusses model capacity, model initialization, training set size and performance on rare words.

\subsection{Model Capacity}
\label{section:capacity}

\begin{figure}
\includegraphics[width=0.5\textwidth]{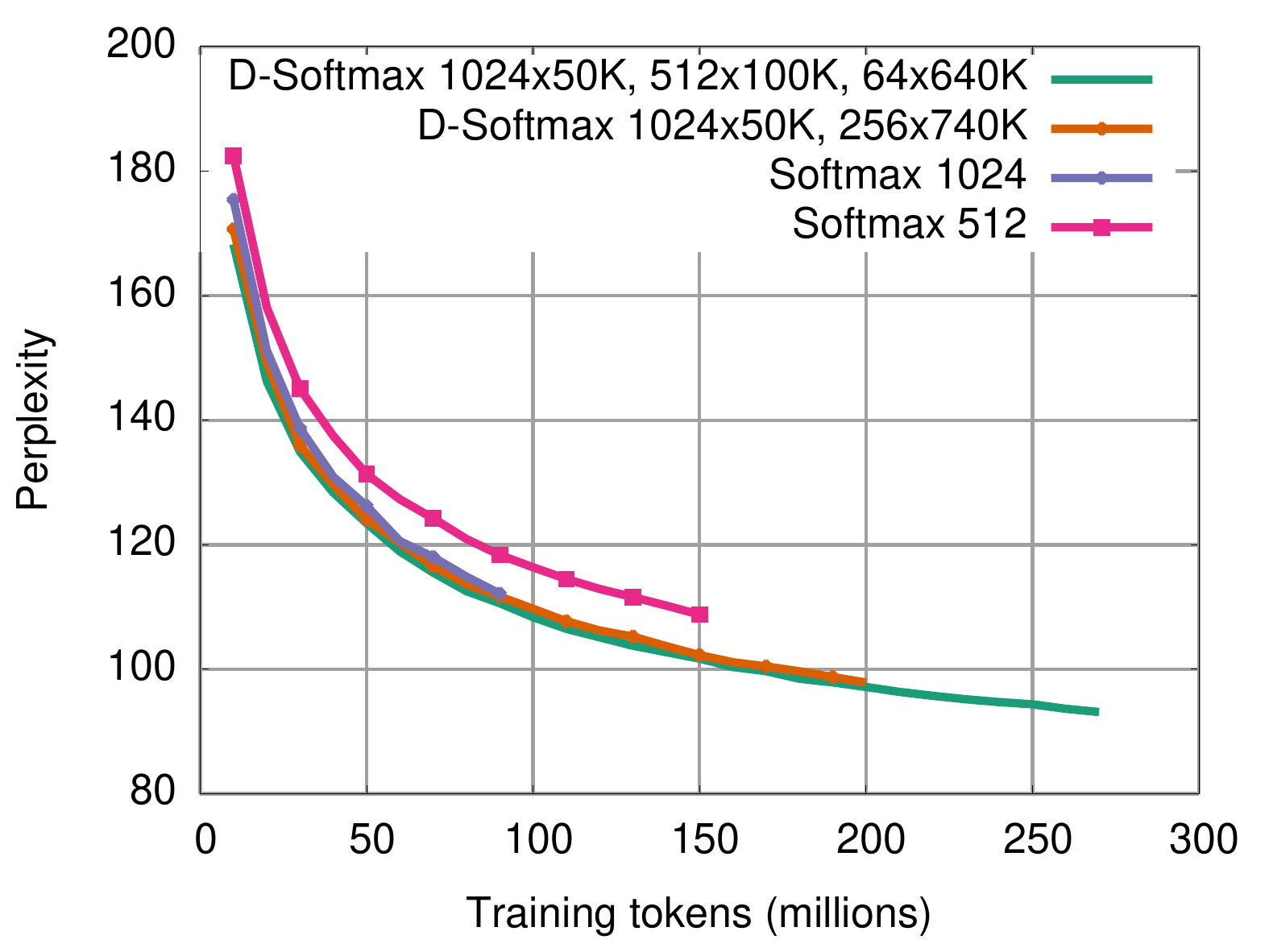}
\caption{Validation perplexity per iteration on billionW for softmax and D-softmax. Softmax uses the same 512 or 1024 units for all words. The first D-Softmax experiment uses 1024 units for the 50K most frequent words, 512 for the next 100K, and 64 units for the rest, the second experiment only considers two frequency bands. All learning curves end after one week.}
\label{figure:capacity_fsm_iter}
\end{figure}

Training neural language models over large corpora highlights that training time, not training data, is the main factor limiting performance. The learning curves on gigaword and billionW indicate that most models are still making progress after one week. Training time has therefore to be taken into account when considering increasing capacity. Figure \ref{figure:capacity_fsm_iter} shows validation perplexity versus the number of iterations for a week of training. This figure indicates that a softmax model with $1024$ hidden units in the last layer could perform better than the $512$-hidden unit model with a longer training horizon. However, in the allocated time, $512$ hidden units yield the best validation performance. D-softmax shows that it is possible to {\it selectively} increase capacity, i.e. to allocate more hidden units to the representation of the most frequent words at the expense of rarer words. This captures most of the benefit of a larger softmax model while staying within a reasonable training budget. 



\subsection{Effect of Initialization}
\label{section:learning_param}

Several techniques for pre-training word embeddings have been recently proposed~\cite{mikolov:2013:corr,lebret:2014:eacl,pennington:2014:emnlp}.
Our experiments use Hellinger PCA~\cite{lebret:2014:eacl}, motivated by its simplicity: it can be computed in a few minutes and only requires an implementation of parallel co-occurrence counting as well as fast randomized PCA. We consider initializing both the input word embeddings and the output matrix from PCA embeddings. 

\begin{figure}
\centering
\includegraphics[width=0.5\textwidth]{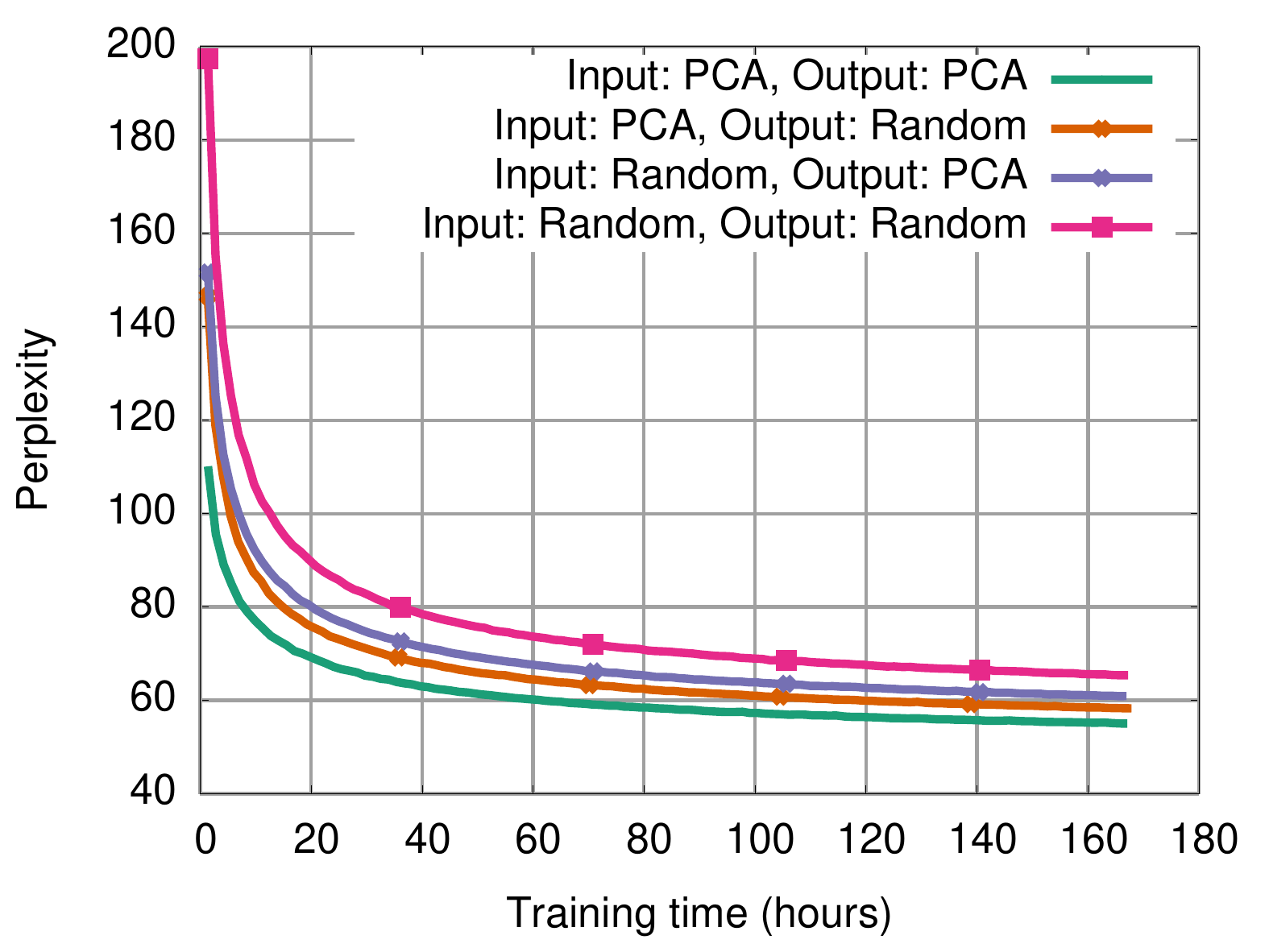}

\caption{Effect of random initialization and with Hellinger PCA on gigaword for softmax.}
\label{figure:pca_learning_curve}
\end{figure}

Figure~\ref{figure:pca_learning_curve} shows that PCA is better than random for initializing both input and output word representations; initializing both from PCA is even better. 
The results show that even after a week of training, the initial conditions still impact the validation perplexity.
This trend is not specific to softmax and similar outcomes have been observed for other strategies.
After a week of training, we observe only for HSM that the random initialization of the output matrix can reach performance comparable to PCA initialization. 

\subsection{Training Set Size}
\label{section:training_set_size}

Large training sets and a fixed training time introduce competition between slower models with more capacity and observing more training data.
This trade-off only applies to iterative SGD optimization and it does not apply to classical count-based models, which visit the training set once and then solve training in closed form.  

\begin{figure}
\centering
\includegraphics[width=0.5\textwidth]{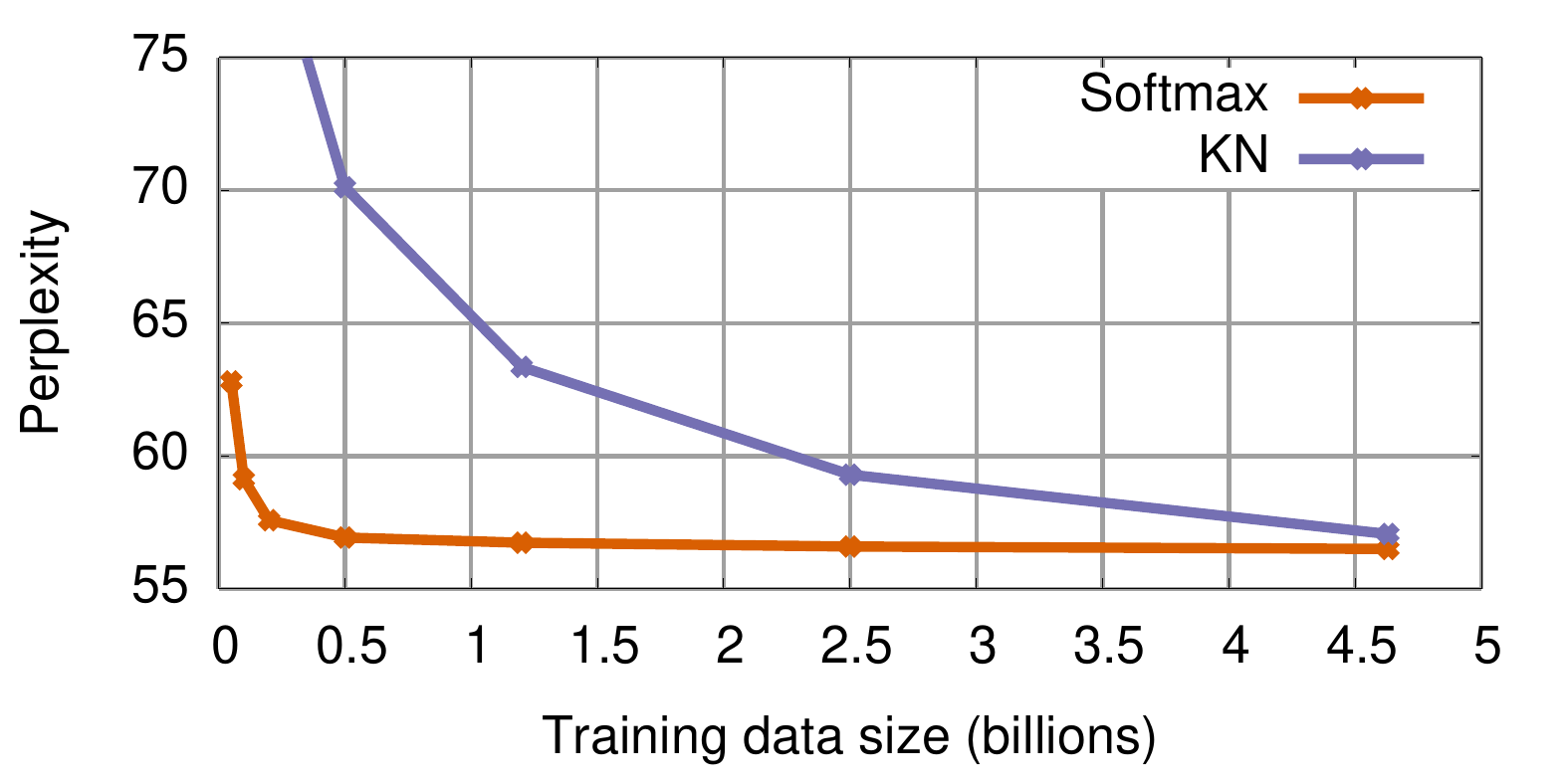}
\caption{Effect of training set size measured on the test set of gigaword for Softmax and Kneser-Ney.}
\label{figure:train_size_gigaword}
\end{figure}

We compare Kneser-Ney and softmax, trained for one week, with gigaword on differently sized subsets of the training data. For each setting we take care to include all data from the smaller subsets. Figure~\ref{figure:train_size_gigaword} shows that the performance of the neural model improves very little on more than 500M tokens. In order to benefit from the full training set we would require a much higher training budget, faster hardware, or parallelization.

\begin{table*}
\centering
\begin{tabular}{l r r r r r}
          & 1-4K & 4-20K & 20-40K & 40-70K & 70-100K \\\hline
Kneser-Ney& 3.48 &7.85  &9.76 &10.76 & 11.57 \\\hline
Softmax   & 3.46 &7.87  &9.76 &11.09 &12.39 \\
D-Softmax & \textbf{3.35} &7.79  &10.13 &12.22 &12.69 \\
Target sampling  & 3.51 &\textbf{7.62} &9.51 &10.81 &12.06 \\
HSM       & 3.49 &7.86 &\textbf{9.38} &\textbf{10.30} &\textbf{11.24} \\
NCE 	  &3.74  &8.48 &10.60  &12.06 &13.37 \\
Weaknorm  &3.46 &7.86 &9.77 &11.12 &12.40 \\
WeaknormSQ &3.46 &7.79 &9.67 &10.98 &12.32 \\
\hline
\end{tabular}
\caption{Test set entropy of various word frequency ranges on gigaword.}
\label{table:overview_entropy}
\end{table*}

Scaling training to large datasets can have a significant impact on perplexity, even when data from the distribution of interest is limited. As an illustration, we adapted a softmax model trained on billionW to Penn Treebank and achieved a perplexity of $96$ - a far better result than with any model we trained from scratch on PTB (cf. Table~\ref{table:overview_results}).

\subsection{Rare Words}
\label{section:rare_words}

How well are neural models performing on rare words?
To answer this question we computed entropy across word frequency bands of the vocabulary for Kneser-Ney and neural models, that is we report entropy for the $4,000$ most frequent words, then the next most frequent $16,000$ words and so on.
Table~\ref{table:overview_entropy} shows that Kneser-Ney is very competitive on rare words, contrary to the common belief that neural models are better on infrequent words.
For frequent words, neural models are on par or better than Kneser-Ney.
This highlights that the two approaches complement each other, as observed in our combination experiments (Table~\ref{table:overview_results}). 

Among the neural strategies, D-Softmax excels on frequent words but performs poorly on rare ones. 
This is because D-Softmax assigns more capacity to frequent words at the expense of rare ones. 
Overall, hierarchical softmax is the best neural technique for rare words since it is very fast. Hierarchical softmax does more iterations than the other techniques and observes the occurrences of every rare words several times.


\section{Conclusions}

This paper presents the first comprehensive analysis of strategies to train large vocabulary neural language models. Large vocabularies are a challenge for neural networks as they need to compute the partition function over the entire vocabulary at each evaluation.

We compared classical softmax to hierarchical softmax, target sampling, noise contrastive estimation and infrequent normalization, commonly referred to as self-normalization. Furthermore, we extend infrequent normalization, or self-normalization, to be a proper estimator of likelihood and we introduce differentiated softmax, a novel variant of softmax which assigns less capacity to rare words in order to reduce computation.

Our results show that methods which are effective on small vocabularies are not necessarily the best on large vocabularies. In our setting, target sampling and noise contrastive estimation failed to outperform the softmax baseline. Overall, differentiated softmax and hierarchical softmax are the best strategies for large vocabularies. Compared to classical Kneser-Ney models, neural models are better at modeling frequent words, but they are less effective for rare words. A combination of the two is therefore very effective. 

From this paper, we conclude that there is still a lot to explore in training from a combination of normalized and unnormalized objectives. We also see parallel training and better rare word modeling as promising future directions.

\toggle{
\section{Acknowledgments}
Do not number the acknowledgment section. Do not include this section when submitting your paper for review.
}

\bibliographystyle{acl}
\bibliography{master}

\begin{thebibliography}{}

\bibitem[\protect\citename{Andreas and Klein}2015]{andreas:2015:naacl}
Jacob Andreas and Dan Klein.
\newblock 2015.
\newblock {When and why are log-linear models self-normalizing?}
\newblock In {\em Proc. of NAACL}.

\bibitem[\protect\citename{Arisoy \bgroup et al.\egroup
  }2012]{arisoy:2012:wfnlm}
Ebru Arisoy, Tara~N. Sainath, Brian Kingsbury, and Bhuvana Ramabhadran.
\newblock 2012.
\newblock {D}eep {N}eural {N}etwork {L}anguage {M}odels.
\newblock In {\em NAACL-HLT Workshop on the Future of Language Modeling for
  HLT}, pages 20--28, Stroudsburg, PA, USA. Association for Computational
  Linguistics.

\bibitem[\protect\citename{Bahdanau \bgroup et al.\egroup
  }2015]{bahdanau:2015:iclr}
Dzmitry Bahdanau, Kyunghyun Cho, and Yoshua Bengio.
\newblock 2015.
\newblock {Neural machine translation by jointly learning to align and
  translate}.
\newblock In {\em Proc. of ICLR}. Association for Computational Linguistics,
  May.

\bibitem[\protect\citename{Bengio \bgroup et al.\egroup
  }2003]{bengio:2003:jmlr}
Yoshua Bengio, R\'{e}jean Ducharme, Pascal Vincent, and Christian Jauvin.
\newblock 2003.
\newblock {A Neural Probabilistic Language Model}.
\newblock {\em Journal of Machine Learning Research}, 3:1137--1155.

\bibitem[\protect\citename{Brown \bgroup et al.\egroup }1992]{brown:1992:cl}
Peter~F. Brown, Peter~V. deSouza, Robert~L. Mercer, Vincent J.~Della Pietra,
  and Jenifer~C. Lai.
\newblock 1992.
\newblock Class-based $n$-gram models of natural language.
\newblock {\em Computational Linguistics}, 18(4):467--479, Dec.

\bibitem[\protect\citename{Chelba \bgroup et al.\egroup
  }2013]{chelba:2013:techreport}
Ciprian Chelba, Tomas Mikolov, Mike Schuster, Qi~Ge, Thorsten Brants, Phillipp
  Koehn, and Tony Robinson.
\newblock 2013.
\newblock One billion word benchmark for measuring progress in statistical
  language modeling.
\newblock Technical report, Google.

\bibitem[\protect\citename{Chopra \bgroup et al.\egroup
  }2015]{chopra:2015:emnlp}
Sumit Chopra, Jason Weston, and Alexander~M. Rush.
\newblock 2015.
\newblock {Tuning as ranking}.
\newblock In {\em Proc. of EMNLP}. Association for Computational Linguistics,
  Sep.

\bibitem[\protect\citename{Devlin \bgroup et al.\egroup }2014]{devlin:2014:acl}
Jacob Devlin, Rabih Zbib, Zhongqiang Huang, Thomas Lamar, Richard Schwartz, ,
  and John Makhoul.
\newblock 2014.
\newblock {Fast and Robust Neural Network Joint Models for Statistical Machine
  Translation}.
\newblock In {\em Proc. of ACL}. Association for Computational Linguistics,
  June.

\bibitem[\protect\citename{Goodman}2001]{goodman:2001:icassp}
Joshua Goodman.
\newblock 2001.
\newblock {Classes for Fast Maximum Entropy Training}.
\newblock In {\em Proc. of ICASSP}.

\bibitem[\protect\citename{Heafield}2011]{heafield:2011:wmt}
Kenneth Heafield.
\newblock 2011.
\newblock {KenLM: Faster and Smaller Language Model Queries}.
\newblock In {\em Workshop on Statistical Machine Translation}, pages 187--197.

\bibitem[\protect\citename{Hyv\"{a}rinen}2010]{gutmann:2010:aistats}
Michael Gutmann~Aapo Hyv\"{a}rinen.
\newblock 2010.
\newblock {Noise-contrastive estimation: A new estimation principle for
  unnormalized statistical models}.
\newblock In {\em Proc. of AISTATS}.

\bibitem[\protect\citename{Jean \bgroup et al.\egroup }2014]{jean:2014:arxiv}
S{\'{e}}bastien Jean, Kyunghyun Cho, Roland Memisevic, and Yoshua Bengio.
\newblock 2014.
\newblock {On Using Very Large Target Vocabulary for Neural Machine
  Translation}.
\newblock {\em CoRR}, abs/1412.2007.

\bibitem[\protect\citename{Le \bgroup et al.\egroup }2012]{le:2012:naacl}
Hai-Son Le, Alexandre Allauzen, and Fran\c{c}ois Yvon.
\newblock 2012.
\newblock {Continuous Space Translation Models with Neural Networks}.
\newblock In {\em Proc. of HLT-NAACL}, pages 39--48, Montr\'{e}al, Canada.
  Association for Computational Linguistics.

\bibitem[\protect\citename{Lebret and Collobert}2014]{lebret:2014:eacl}
Remi Lebret and Ronan Collobert.
\newblock 2014.
\newblock {Word Embeddings through Hellinger PCA}.
\newblock In {\em Proc. of EACL}.

\bibitem[\protect\citename{LeCun \bgroup et al.\egroup
  }1998]{lecun:1998:tricks}
Yann LeCun, Leon Bottou, Genevieve Orr, and Klaus-Robert Mueller.
\newblock 1998.
\newblock {Efficient BackProp}.
\newblock In Genevieve Orr and Klaus-Robert Muller, editors, {\em {Neural
  Networks: Tricks of the trade}}. Springer.

\bibitem[\protect\citename{Marcus \bgroup et al.\egroup }1993]{marcus:1993:cl}
Mitchell~P. Marcus, Mary~Ann Marcinkiewicz, and Beatrice Santorini.
\newblock 1993.
\newblock {Building a Large Annotated Corpus of English: The Penn Treebank}.
\newblock {\em Computational Linguistics}, 19(2):314--330, Jun.

\bibitem[\protect\citename{Mikolov \bgroup et al.\egroup
  }2010]{mikolov:2010:interspeech}
Tom{\'a}\v{s} Mikolov, Karafi{\'a}t Martin, Luk{\'a}\v{s} Burget, Jan
  Cernock{\'y}, and Sanjeev Khudanpur.
\newblock 2010.
\newblock {Recurrent Neural Network based Language Model}.
\newblock In {\em Proc. of INTERSPEECH}, pages 1045--1048.

\bibitem[\protect\citename{Mikolov \bgroup et al.\egroup
  }2011a]{mikolov:2011:interspeech}
Tomas Mikolov, Anoop Deoras, Stefan Kombrink, Lukas Burget, and Jan~Honza
  Cernocky.
\newblock 2011a.
\newblock {Empirical Evaluation and Combination of Advanced Language Modeling
  Techniques}.
\newblock In {\em Interspeech}.

\bibitem[\protect\citename{Mikolov \bgroup et al.\egroup
  }2011b]{mikolov:2011:icassp}
Tom{\'a}\v{s} Mikolov, Stefan Kombrink, Luk{\'a}\v{s} Burget, Jan Cernock{\'y},
  and Sanjeev Khudanpur.
\newblock 2011b.
\newblock {Extensions of Recurrent Neural Network Language Model}.
\newblock In {\em Proc. of ICASSP}, pages 5528--5531.

\bibitem[\protect\citename{Mikolov \bgroup et al.\egroup
  }2013]{mikolov:2013:corr}
Tom{\'a}\v{s} Mikolov, Kai Chen, Greg Corrado, and Jeffrey Dean.
\newblock 2013.
\newblock {Efficient Estimation of Word Representations in Vector Space}.
\newblock {\em CoRR}, abs/1301.3781.

\bibitem[\protect\citename{Mnih and Hinton}2010]{mnih+hinton:2008:nips}
Andriy Mnih and Geoffrey~E. Hinton.
\newblock 2010.
\newblock {A Scalable Hierarchical Distributed Language Model}.
\newblock In {\em Proc. of NIPS}.

\bibitem[\protect\citename{Mnih and Teh}2012]{mnih:2012:icml}
Andriy Mnih and Yee~Whye Teh.
\newblock 2012.
\newblock {A fast and simple algorithm for training neural probabilistic
  language models}.
\newblock In {\em Proc. of ICML}.

\bibitem[\protect\citename{Morin and Bengio}2005]{morin:2005:aistats}
Frederic Morin and Yoshua Bengio.
\newblock 2005.
\newblock {Hierarchical Probabilistic Neural Network Language Model}.
\newblock In {\em Proc. of AISTATS}.

\bibitem[\protect\citename{Parker \bgroup et al.\egroup
  }2011]{parker:2011:techreport}
Robert Parker, David Graff, Junbo Kong, Ke~Chen, and Kazuaki Maeda.
\newblock 2011.
\newblock {English Gigaword Fifth Edition}.
\newblock Technical report, Linguistic Data Consortium.

\bibitem[\protect\citename{Pennington \bgroup et al.\egroup
  }2014]{pennington:2014:emnlp}
Jeffrey Pennington, Richard Socher, and Christopher~D Manning.
\newblock 2014.
\newblock Glove: Global vectors for word representation.
\newblock In {\em Proceedings of the Empiricial Methods in Natural Language
  Processing}.

\bibitem[\protect\citename{Schwenk \bgroup et al.\egroup
  }2012]{schwenk:2012:wfnlm}
Holger Schwenk, Anthony Rousseau, and Mohammed Attik.
\newblock 2012.
\newblock {Large, Pruned or Continuous Space Language Models on a {GPU} for
  Statistical Machine Translation}.
\newblock In {\em NAACL-HLT Workshop on the Future of Language Modeling for
  HLT}, pages 11--19. Association for Computational Linguistics.

\bibitem[\protect\citename{Sordoni \bgroup et al.\egroup
  }2015]{sordoni:2015:naacl}
Alessandro Sordoni, Michel Galley, Michael Auli, Chris Brockett, Yangfeng Ji,
  Margaret Mitchell, Jian-Yun Nie1, Jianfeng Gao, and Bill Dolan.
\newblock 2015.
\newblock {A Neural Network Approach to Context-Sensitive Generation of
  Conversational Responses}.
\newblock In {\em Proc. of NAACL}. Association for Computational Linguistics,
  May.

\bibitem[\protect\citename{Sutskever \bgroup et al.\egroup
  }2014]{sutskever:2014:nips}
Ilya Sutskever, Oriol Vinyals, and Quoc Le.
\newblock 2014.
\newblock {Sequence to Sequence Learning with Neural Networks}.
\newblock In {\em Proc. of NIPS}.

\bibitem[\protect\citename{Vaswani \bgroup et al.\egroup
  }2013]{vaswani:2013:emnlp}
Ashish Vaswani, Yinggong Zhao, Victoria Fossum, and David Chiang.
\newblock 2013.
\newblock {Decoding with Large-scale Neural Language Models improves
  Translation}.
\newblock In {\em Proc. of EMNLP}. Association for Computational Linguistics,
  October.

\bibitem[\protect\citename{Vijayanarasimhan \bgroup et al.\egroup
  }2014]{vijayanarasimhan:2014:arxiv}
Sudheendra Vijayanarasimhan, Jonathon Shlens, Rajat Monga, and Jay Yagnik.
\newblock 2014.
\newblock Deep networks with large output spaces.
\newblock {\em CoRR}, abs/1412.7479.

\end{thebibliography}

\end{document}